\documentclass[10pt,twocolumn,letterpaper]{article}

\usepackage[pagenumbers]{iccv} % To force page numbers, e.g. for an arXiv version

\usepackage{mdwmath}
\usepackage{eqparbox}
\usepackage{epsfig}
\usepackage{graphicx}
\usepackage{amsmath}
\usepackage{amssymb}
\usepackage{pifont}
\usepackage{url}            % simple URL typesetting
\usepackage{booktabs}       % professional-quality tables
\usepackage{tabu}           % tabular
\usepackage{multirow}
\usepackage{graphicx}
\usepackage{algorithm}
\usepackage{algorithmicx}
\usepackage{algpseudocode}
\usepackage{amsmath,amssymb}
\usepackage{array}

\newcommand{\cmark}{\ding{51}}%
\newcommand{\xmark}{\ding{55}}%
\usepackage{cite}

\usepackage{makecell}
\usepackage{tabularx}
\usepackage{pifont}
\usepackage{multicol}
\usepackage{array}
\usepackage{adjustbox}

\usepackage{url}            % simple URL typesetting
\usepackage{booktabs}       % professional-quality tables
\usepackage{amsfonts}       % blackboard math symbols
\usepackage{nicefrac}       % compact symbols for 1/2, etc.
\usepackage{microtype}      % microtypography
\usepackage{graphicx}
\usepackage{amsmath}
\usepackage{bm}
\usepackage{multirow}
\usepackage{tabu}
\usepackage{siunitx}
\usepackage{adjustbox}
\usepackage{subcaption}
\usepackage{siunitx}
\usepackage{colortbl}
\usepackage{color}
\usepackage{pifont}
\definecolor{Gray}{gray}{0.9}
\definecolor{Lightorange}{RGB}{255,214,169}
\definecolor{Cyan}{rgb}{0.88,1,1}
\definecolor{reminder}{RGB}{255,0,0}

\usepackage{tabu}           % tabular
\usepackage{multirow}
\usepackage{booktabs}
\usepackage{makecell}
\usepackage{pifont}
\usepackage{longtable}

\newcommand{\paragrapha}[2][3pt]{\vspace{#1}\noindent\textbf{#2}}

\newcolumntype{x}[1]{>{\centering\arraybackslashå}p{#1pt}}
\newlength\savewidth

\newcommand{\PreserveBackslash}[1]{\let\temp=\\#1\let\\=\temp}
\newcolumntype{C}[1]{>{\PreserveBackslash\centering}p{#1}}
\newcolumntype{L}[1]{>{\PreserveBackslash\raggedright}p{#1}}

\usepackage{url}

\definecolor{iccvblue}{rgb}{0.21,0.49,0.74}
\usepackage[pagebackref,breaklinks,colorlinks,allcolors=iccvblue]{hyperref}

\def\paperID{xxxx} % *** Enter the Paper ID here
\def\confName{ICCV}
\def\confYear{2025}

\title{SparseMM: Head Sparsity Emerges from Visual Concept Responses in MLLMs}

\author{Jiahui Wang\textsuperscript{1}\thanks{Authors contributed equally to this research.~~\textsuperscript{\dag}Corresponding author.},~~Zuyan Liu\textsuperscript{1,2}\footnotemark[1],~~Yongming Rao\textsuperscript{2,1},~~Jiwen Lu\textsuperscript{1}$^{\dagger}$ \\
\textsuperscript{1}~Tsinghua University~~\textsuperscript{2}~Tencent Hunyuan X \\
}

\begin{document}
\maketitle
\begin{abstract}
Multimodal Large Language Models (MLLMs) are commonly derived by extending pre-trained Large Language Models (LLMs) with visual capabilities. In this work, we investigate how MLLMs process visual inputs by analyzing their attention mechanisms. We reveal a surprising sparsity phenomenon: only a small subset (approximately less than 5\%) of attention heads in LLMs actively contribute to visual understanding, termed \textbf{visual heads}. To identify these heads efficiently, we design a training-free framework that quantifies head-level visual relevance through targeted response analysis. Building on this discovery, we introduce \textbf{SparseMM}, a KV-Cache optimization strategy that allocates asymmetric computation budgets to heads in LLMs based on their visual scores, leveraging the sparity of visual heads for accelerating the inference of MLLMs. Compared with prior KV-Cache acceleration methods that ignore the particularity of visual, SparseMM prioritizes stress and retaining visual semantics during decoding. Extensive evaluations across mainstream multimodal benchmarks demonstrate that SparseMM achieves superior accuracy-efficiency trade-offs. Notably, SparseMM delivers 1.38× real-time acceleration and 52\% memory reduction during generation while maintaining performance parity on efficiency test. Our project is open sourced at \url{https://github.com/CR400AF-A/SparseMM}.
\end{abstract}    
\section{Introduction}
\label{sec:intro}

\begin{figure}[t]
\centering
\includegraphics[width=0.5\textwidth]{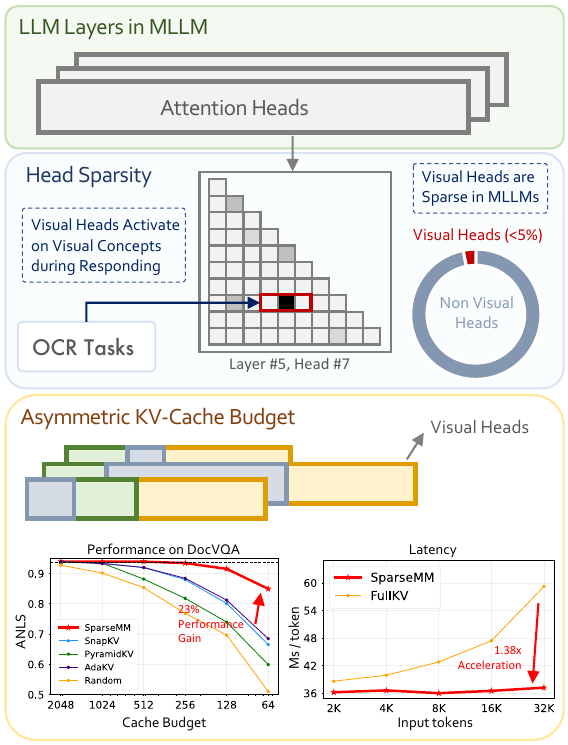} \vspace{-10pt}
\caption{\textbf{Head Sparsity Emerges from Visual Concept Responses.} We observe the visual-relevant heads are sparse in various MLLMs. Based on this observation, we devise a  KV-Cache optimization strategy that allocates asymmetric budgets to LLM heads  based on their importance for visual tokens, achieving better trade-off under limited computational resources. }
\label{fig:teaser}  \vspace{-15pt}
\end{figure}

Autoregressive large language models (LLMs)~\citep{GPT4o,GPT35,reid2024gemini, chen2024intern25, qwen2.5} have revolutionized artificial intelligence with their exceptional instruction-following capabilities and expansive knowledge repositories. Building upon this foundation, researchers have extended LLMs to multimodal domains, particularly in vision-language integration, creating multimodal large language models (MLLMs)~\citep{li2023blip2,alayrac2022flamingo,liu2024llava15,chen2024internvl,ye2024mplug,Qwen2.5-VL} that process both textual and visual inputs.  Current approaches typically augment pre-trained LLMs by incorporating visual encoders (e.g., CLIP~\citep{radford2021clip} or SigLIP~\citep{zhai2023siglip}) paired with lightweight adapters to project visual features into the language model's hidden space. While these architectures demonstrate remarkable multimodal reasoning abilities, how LLMs fundamentally acquire visual comprehension during supervised fine-tuning remains poorly understood.  This knowledge gap constrains our ability to recognize cross-modal alignment and risks undervaluing visual semantics during multi-modal relevant tasks and applications, which may potentially leading to suboptimal architecture designs and inefficient computational resource allocation.

To this end, we present the first systematic investigation into how visual concepts are processed within LLMs.    Through rigorous analysis of attention mechanisms, we uncover a critical phenomenon that only a small subset of attention heads (termed \textbf{visual heads}) drive visual content understanding, while the majority remain text-specialized. Specifically, our experiments reveal two critical properties of visual heads: (1) \textit{Sparsity}: Less than 5\% of attention heads are intrinsically visual-active across layers, even in models trained with extensive multimodal data;    (2) \textit{Universality}: Visual heads emerge consistently across diverse LLM architectures(e.g., Vicuna~\citep{chiang2023vicuna} and Qwen2~\citep{qwen2vl}) and generalize to multiple attention paradigms such as multi-head attention (MHA)~\citep{vaswani2017attention} and grouped query attention (GQA)~\citep{ainslie2023gqa}.

To systematically identify these visual heads, we propose a training-free framework that quantifies the visual relevance of attention heads through targeted cross-modal response analysis. Specifically, our approach leverages OCR as an anchor task to establish precise correspondence between text outputs and visual inputs: for each generated word, we trace its activation back to spatially aligned image patches, enabling direct measurement of how specific attention heads mediate visual-text alignment.  By analyzing and recoding the attention score of all the attention heads across a certain amount of samples, we compute visual scores that rank heads by their visual responsiveness. Crucially, while our identification mechanism relies on OCR’s granular spatial grounding, we demonstrate that the detected visual heads exhibit task-agnostic generalizability—they remain dominant in diverse vision-language tasks including object recognition and scene understanding.

Building on these insights, to demonstrate the effectiveness of visual heads on practical multi-modal tasks, we introduce \textbf{SparseMM}, a KV-Cache optimization framework that exploits visual head sparsity to achieve accelerated inference. As multimodal inputs grow in complexity—spanning multi-turn dialogues~\citep{yao2024minicpm,li2024llavaov,li2024llavainterleave}, high-resolution interleaved images~\citep{xu2024llavauhd,dong2024internlm}, and dense video/3D sequences~\citep{fei2024videoccam,lin2023videollava,hong20233d}—the computational overhead of maintaining full KV-Caches becomes prohibitive. Existing compression methods, however, treat all attention heads uniformly, disregarding the critical role of sparse visual heads in encoding visual semantics.

SparseMM addresses this by asymmetrically allocating KV-Cache budgets: visual heads receive prioritized retention based on their precomputed visual scores, while non-visual heads undergo aggressive compression via a hybrid strategy combining 1) \textit{Score-Preferred Cache} (allocating cache budget based on visual head scores), 2) \textit{Uniform-Based Cache} (preserving minimal budget for all the heads), and 3) \textit{Local Window Cache} (preserving cache budget for recent tokens). This mixed approach ensures better accuracy-efficiency trade-offs, such that visual heads retain more computational cost while other heads are dynamically throttled.

Extensive experimental results demonstrate that SparseMM outperforms other strong baselines across multiple datasets, including DocVQA~\citep{mathew2021docvqa}, OCRBench~\citep{liu2023ocrbench}, TextVQA~\citep{singh2019textvqa}, MMBench~\citep{liu2023mmbench}, etc. For instance, on DocVQA, LLaVA-NeXT-Vicuna-7B~\citep{liu2024llavanext} achieves the same level of accuracy while using only 20\% of the cache, and Qwen2-VL-7B-Instruct~\citep{qwen2vl} achieves equivalent performance with just 5.3\% of the cache. These findings suggest that our method effectively captures visual information while compressing redundancies. Furthermore, the reduction in cache requirements enables our method to achieve lower decoding latency and peak memory usage. For example, LLaVA-NeXT-Mistral-7B~\citep{liu2024llavanext} maintains nearly constant decoding latency with 32K input tokens, resulting in almost a 50\% acceleration compared to the full model, and reduces memory usage by 5GB.
\section{Related Works}

\paragrapha{Architectures in MLLMs. }
The predominant architecture for Multi-Modal Large Language Models (MLLMs) consists of three key components: a visual encoder, an adapter, and a LLM. By leveraging alignment training techniques and subsequent fine-tuning, this integrated framework has achieved remarkable performance on various multi-modal understanding tasks~\citep{GPT4V,li2024llavaov,lin2023vila,agrawal2024pixtral,liu2024oryx,liu2025ola,dong2024insight}. In typical implementations, the visual encoder is realized using models like CLIP~\citep{radford2021clip} or SigLIP~\citep{zhai2023siglip}, which are adept at extracting rich visual representations. The adapter component serves as an intermediary, bridging the gap between the visual features and the language domain; it is often instantiated as a multi-layer perceptron (MLP) or through more complex structures~\citep{lu2024ovis}. For the LLM part, architectures such as LLaMA~\citep{touvron2023llama,touvron2023llama2} and Qwen~\citep{qwen2} series are commonly employed. Notably, previous LLMs such as LLaMA2~\citep{touvron2023llama2} utilizes a multi-head attention (MHA) mechanism, while more recent LLMs such as LLaMA3~\citep{dubey2024llama3} incorporate a grouped query attention (GQA)~\citep{ainslie2023gqa} design. The GQA approach aggregates multiple queries into a single group that corresponds to one key and one value, thereby substantially reducing memory usage without compromising model performance. In this paper, we observe a universal phenomenon of visual head in MLLMs across LLM architectures and attention implementations. The applications are demonstrated to be effective across the abundant MLLM series.

\begin{figure*}[t]
\centering
\includegraphics[width=\textwidth]{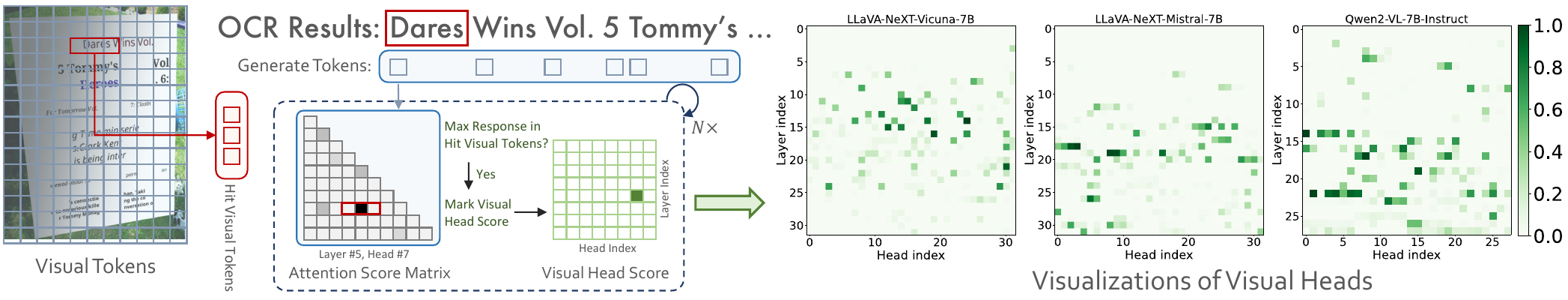}
\caption{\textbf{Visual Heads are Sparse in MLLMs. } We use OCR tasks to obtain visual scores for all heads. Upon visualizing these scores, we discovered that high-scoring heads, which we refer to as \textbf{visual heads}, are quite sparse within the MLLM, comprising only about 5\%. The majority of heads have very low scores, indicating that most heads in LLMs do not focus on visual information.}
\label{fig:visualhead} \vspace{-10pt}
\end{figure*}

\paragrapha{Model Acceleration in MLLMs. }
With the rapid growth of model size and input sequence length, model acceleration has become an urgent research focus in both language and multi-modal domains. In the context of Large Language Models (LLMs), significant efforts have been dedicated to optimizing the prompt encoding phase through efficient compression of the KV-Cache. For instance, StreamingLLM~\citep{xiao2023streamingllm} identifies attention sinks to stabilize long-context inference, while H2O~\citep{zhang2023h2o} introduces a token-level importance scoring mechanism for adaptive KV-Cache eviction. Subsequent works, such as SnapKV~\citep{li2024snapkv}, PyramidKV~\citep{cai2024pyramidkv}, and AdaKV~\citep{feng2024adakv}, further refine the KV-Cache selection strategy by incorporating spatial-temporal redundancy reduction, hierarchical token retention, or dynamic eviction policies. However, these methods, primarily designed for text-only inputs, face limitations when applied to multi-modal scenarios. In MLLMs, acceleration challenges are exacerbated by the increasing complexity of multi-modal prompts (e.g., high-resolution images, videos) and cross-modal fusion mechanisms. Recent attempts, such as FastV~\citep{chen2024fastv}, accelerate inference via layer-wise pruning of redundant visual tokens. ElasticCache~\citep{liu2024elastic}, optimizes KV-Cache management during the generation phase. Despite these advances, head-wise acceleration strategies, particularly those targeting modality-specific attention heads, remain underexplored.  In this work, we address this gap by proposing a systematic framework based on our findings about visual heads, enabling efficient deployment of MLLMs in resource-constrained environments.
\section{Visual Heads are Sparse in MLLMs}

In this section, we present our exploration of head sparsity in multi-modal large language models. To start with, we provide the preliminaries on the relations from LLMs to visual instruction tuning. Then we describe our approach for identifying sparse visual heads in MLLMs in Sec.~\ref{3_method_chasing}, then introduce the deployment of visual heads in model acceleration in Sec.~\ref{3_method_sparsemm}.

\begin{algorithm}[t]
\renewcommand{\algorithmicrequire}{\textbf{Input:}}
\renewcommand{\algorithmicensure}{\textbf{Output:}}

\caption{Chasing Visual Heads in MLLMs}\label{alg:algo}
\begin{algorithmic}[1]
\Require
\Statex ocr\_text\_bbox\_pair = List[(text, bbox)]
\Statex output\_token = $\{y_i\}_{i=1}^{N}$
\Statex image\_shape, feature\_map

\Ensure Matrix $S$ representing scores of heads in LLMs
\For {$i$ = 1 to $N$}
\State bbox = match($y_i$, ocr\_text\_bbox\_pair)
\State patch\_idx = match(bbox, image\_shape, feature\_map)  %\# map the bbox to nearest patches.
\State image\_tokens = find(patch\_idx, feature\_map) %\# according to the iamges's shape.

    \For {(layer, head)}
        % \For {head}
            \State index = argmax($A_\text{head}^\text{layer}$)
            \If {index \textbf{in} image\_tokens}
                \State $S_\text{head}^\text{layer}$ += $\frac{1}{\#\text{image\_tokens}}$
            \EndIf
        % \EndFor
    \EndFor
\EndFor

\State \textbf{Return} $S$
\end{algorithmic}
\end{algorithm}

\subsection{What is Learned during Visual Instruct Tuning}
% intro, we aim to figure out what is learned from LLM -> MLLM
Extending a Large Language Model to a Multimodal Large Language Model is achieved by integrating a visual encoder $E$, an adapter $H$, and the LLM $p_{\theta}$. The original LLM is trained solely on textual tasks to model the distribution of text sequences as $p_{\theta}(\mathbf{x})=\prod_{i=1}^{N}p_{\theta}(\mathbf{x}_i|\mathbf{x}_{<i})$, where $\{\mathbf{x_i}\}_{i=1}^N$. The visual encoder, typically based on architectures such as CLIP~\citep{radford2021clip} or SigLIP~\citep{zhai2023siglip}, is responsible for extracting visual features from images. An adapter $H$ is then utilized to project these visual features into the semantic space, culminating in a multimodal model that can be formally represented as follows:
\begin{equation}
    p_{\theta}(\mathbf{x})=\prod_{i=1}^{N}p_{\theta}(\mathbf{x}_i|\mathbf{x}_{<i}, \mathbf{v}), \mathbf{v}=H(E(\text{Image}))
\end{equation}
In order to enable the LLM to comprehend and process visual information, a pre-alignment phase is conducted, followed by a visual instruction fine-tuning stage. The objective during these stages is to minimize the cross-entropy loss between the generated textual output and the ground truth as follows:
\begin{equation}
\mathcal{L} = -\frac{1}{N-P}\sum_{i=P+1}^{N}\text{log}p_{\theta}(\mathbf{x}_i|\mathbf{x}_{<i}, \mathbf{v})
\end{equation}
where $P$ is the length of input tokens.

Although the resulting MLLM shows outstanding performance in various tasks, the precise modifications that occur during the transition from LLM to MLLM, rendering the model capable of understanding visual information, are not sufficiently understood. However, we found that some attention heads within the MLLM have learned to focus on visual information during the visual instruction finetuning process. We refer to these heads as \textbf{visual heads}.

\subsection{Chasing Visual Heads in MLLMs} \label{3_method_chasing}
% how to find visual head, our method/some results
To investigate how attention heads within the MLLM attend to visual elements and to identify the specific visual head, we introduce an OCR-based method and define the visual score. As in Alg.~\ref{alg:algo} and Fig.~\ref{fig:visualhead}, for a given text instruction and OCR image as input $X$, the MLLM is tasked with generating the OCR output. For each output token $y_i$, we first determine its corresponding region within the image based on (text, bbox) pair. Based on this region, we then identify the associated image tokens denoted $I_{y_i}$ in the input sequence:
\begin{equation} 
\text{Visual Score for Head}~h = \frac{1}{N}\sum_{i=1}^{N} \frac{\mathbb{I}_{hit(y_i, A_h)}}{\#\text{image\_tokens}}
\end{equation}
where
\begin{equation}
\text{hit}(y_i,A_h)=
\begin{cases}
1, & \text{argmax}(A_h) \in I_{y_i}  \\
0, & \text{else}
\end{cases}
\end{equation}
Subsequently, we iterate over all attention heads. For any given head $h$, if the token that receives the highest attention in this head's attention matrix $A_h$ belongs to the set of identified image tokens, a ``hit" is recorded for that head and its score is incremented by the inverse of the number of image tokens. This means a smaller (more precise) region yields a higher score, because they are harder to capture.

Finally, we aggregate the scores from all heads across 1,000 OCR images from the Synthdog dataset~\citep{kim2021donut}. These scores are then normalized to produce a score matrix, the visualization of which is presented in Fig.~\ref{fig:visualhead}.

\subsection{Exploring Head Sparsity for Acceleration} \label{3_method_sparsemm}
% method for acceleration

In multi-modal models, visual tokens comprise a significant portion of the input sequence, and each token necessitates its own key-value (KV) cache. This requirement leads to a substantial and often prohibitive increase in computational cost and memory consumption. However, previous analyses have shown that not every attention head relies highly on visual information. This finding motivates a natural idea: allocate varying KV-cache budgets to different attention heads in proportion to their visual attention scores, thereby balancing efficiency with overall performance. 

\begin{figure}[t]
\centering
\includegraphics[width=0.5\textwidth]{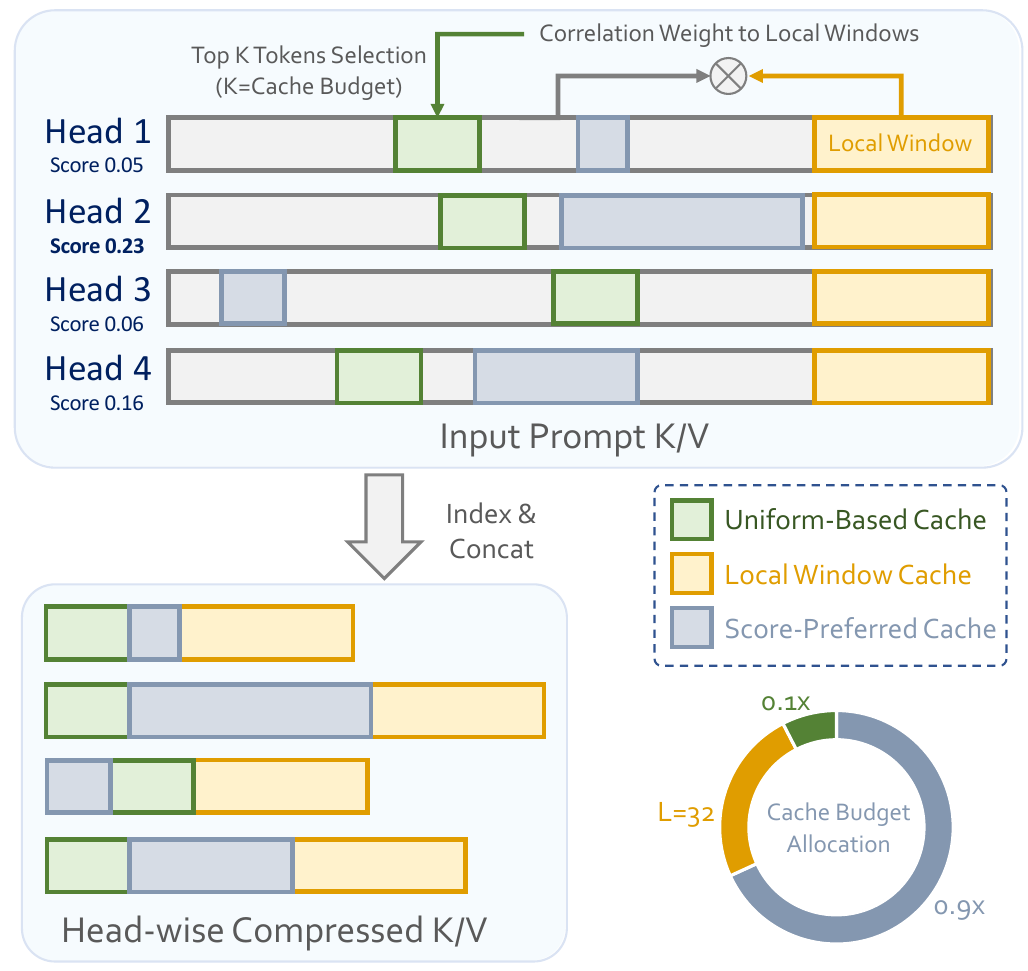}
\caption{\textbf{SparseMM for MLLM Acceleration. } The KV Cache budget for each head is composed of three parts: \textbf{Local Window Cache}, \textbf{Uniform-Based Cache}, and \textbf{Score-Preferred Cache}. The top-K KV caches are selected based on attention scores.}
\label{fig:sparsemm} \vspace{-10pt}
\end{figure}

In this subsection, we describe \textbf{SparseMM} for allocating each head's cache budget, as illustrated in Fig.~\ref{fig:sparsemm}. For a typical multi-modal model with \(L\) layers and \(H\) heads per layer, we can obtain a visual attention score matrix \(Score_{L \times H}\) as detailed in Sec.~\ref{3_method_chasing}. In an ideal setting, the cache allocation would be determined exclusively by the values in \(Score_{L \times H}\). However, inspired by AdaKV~\citep{feng2024adakv}, and to account for locality and to ensure that every head maintains a minimum level of budget, we introduce a three-part allocation mechanism:

\paragrapha{1) Local Window Cache}: Each head is first allocated a fixed, predetermined cache size for the nearest neighbor window. We denote this window size by $w$, with a default value of $32$. Thus, the total cache allocated for all heads in this step is $N\cdot w$

\paragrapha{2) Uniform-Based Cache}: Denote the total budget by \( B \). From the remaining budget, 
\begin{equation}
B_{\text{remain1}} = B - N \cdot w
\end{equation}
a fixed ratio, denoted by $\rho\in (0,1)$(with a default value of $0.1$), of this remainder is uniformly allocated to each head. That is, each head receives an additional baseline cache of 
\begin{equation}
r = \frac{\rho \cdot \left(B - N \cdot w\right)}{N}
\end{equation}

\paragrapha{3) Score-Preferred Cache}: The remaining budget after the uniform allocation,
\begin{equation}
B_{\text{remain2}} = B - N\cdot w - \rho\left(B - N\cdot w\right)
\end{equation}
is then distributed among the heads in proportion to their corresponding visual attention scores. We denote by $s_{ij}$ the element in the $i$th row and $j$th column of the matrix \(Score_{L \times H}\), which represents the visual attention score of the $j$th head in the $i$th layer. Then, the score-based cache allocated to head $(i,j)$ is given by
\begin{equation}
b_{ij}^{\text{score}} = B_{\text{remain2}} \cdot \frac{s_{ij}}{\displaystyle \sum_{i=1}^{L}\sum_{j=1}^{H} s_{ij}}
\end{equation}

Summing the contributions from each of the three parts, the final cache allocation for head $(i,j)$ is expressed as
\begin{equation}
b_{ij} = w + r + b_{ij}^{score}
\end{equation}

Once each head establishes its budget, the most salient KV Caches are identified by ranking the attention scores. Inspired by the approach presented in SnapKV~\citep{li2024snapkv}, which employs an observation window at the end of the prompt, we restrict our attention computation to only the final observation window of size 32. Assume we have Query\_States Q and Key\_States K, then we compute local window attention as follows: 

\begin{equation}
A = \text{softmax}(\frac{\mathbf{Q}_{\text{loc}}\, \mathbf{K}_{\text{all}}^{\top}}{\sqrt{d}} + M)
\end{equation}
where 
\begin{equation}
\mathbf{Q}_{\text{loc}} = \mathbf{Q}[:, :, L-w : L, :], \quad  \mathbf{K}_{\text{all}} = \mathbf{K}
\end{equation}
\begin{equation}
M_{i,j} =
\begin{cases}
0, & \text{if } j \leq i \\
-\infty, & \text{if } j > i
\end{cases}
\end{equation}

This strategy effectively reduces the computational complexity from $O(N^2)$ to $O(N \times L)$, where $L=32$, thereby decreasing the runtime during the prefilling stage. To evaluate the attention for keys outside the local window, we compute the average attention weight:
\begin{equation}
\bar{A}_j = \frac{1}{w} \sum_{i=1}^{w} \hat{A}_{i,j}, \quad \text{for } j \in \{0, \dots, L-w-1\}
\end{equation}

Ultimately, we select the top $K$ KV Caches based on the computed attention scores, where the value of $K$ is given by
\begin{equation}
K = r + b_{ij}
\end{equation}

Our three-part allocation mechanism leverages head sparsity to significantly reduce the computational and memory overhead in multi-modal models. It ensures that each head receives a guaranteed minimum cache allocation through both the nearest neighbor and uniform baseline allocations. The remaining cache is then adaptively distributed based on the visual attention scores, thereby achieving an efficient balance between computational efficiency and overall model performance.

\section{Experiments}
We conduct extensive experiments to validate the effectiveness of Visual Head. We first introduce our experimental settings in Section \ref{subsec:exp_setting}. Then we present a comparison with state-of-the-art KV Cache Compression method, demonstrating that our method maintains strong performance in Section \ref{subsec:exp_result} while maintaining computational efficiency in Section \ref{subsec:exp_efficiency}. Moreover, we provide some analytical experiments to illustrate the importance of visual head in Section \ref{subsec:exp_analysis}.

\subsection{Experimental Settings}\label{subsec:exp_setting}

\begin{figure*}[t]
  \centering
  \includegraphics[width=\linewidth]{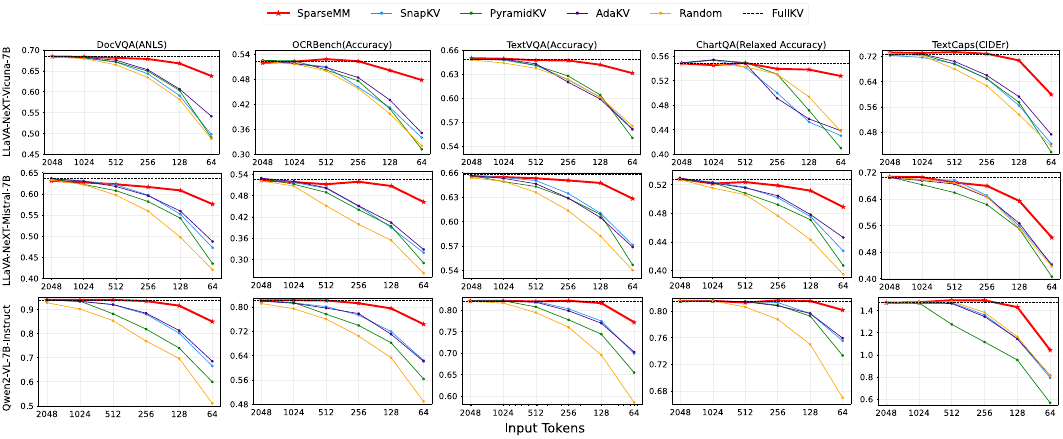}\vspace{-5pt}
  \caption{\textbf{Main Results on Multi-Modal Benchmarks.} We evaluate SparseMM and other baselines on several multimodal benchmarks, and conduct experiments on a series of backbones. Our SparseMM consistently outperforms the other baselines.}
  \label{fig:main_result}\vspace{-10pt}
\end{figure*}

\paragrapha{Models.}
We employ three multi-modal models: LLaVA-NeXT-Vicuna-7B~\citep{liu2024llavanext}, LLaVA-NeXT-Mistral-7B~\citep{liu2024llavanext}, and Qwen2-VL-7B~\citep{qwen2vl}. LLaVA-NeXT-Vicuna-7B is derived from Vicuna-7B~\citep{chiang2023vicuna}, a model based on Multi-Head Attention (MHA)~\citep{vaswani2017attention}, and comprises 32 layers with 32 attention heads per layer. In contrast, LLaVA-NeXT-Mistral-7B is built upon Mistral-7B~\citep{jiang2024mixtral} and utilizes Grouped-Query Attention (GQA)~\citep{ainslie2023gqa}. This model features 32 layers, with each layer consisting of 32 query heads and 8 key-value heads. Similarly, Qwen2-VL-7B~\citep{qwen2vl} is based on Qwen2, another GQA model, and is composed of 28 layers, with each layer containing 28 query heads and 4 key-value heads.

\paragrapha{Baselines. }
We adopt SnapKV~\citep{li2024snapkv}, PyramidKV~\citep{cai2024pyramidkv}, and AdaKV~\citep{feng2024adakv} as our baseline methods, as they represent the latest and state-of-the-art in KV Cache compression. SnapKV~\citep{li2024snapkv} utilizes an ``observation window" mechanism to identify and preserve the most critical KV caches. PyramidKV~\citep{cai2024pyramidkv} implements a hierarchical allocation strategy that distributes the KV cache budget in a pyramidal manner. More budget is allocated to lower layers with dispersed attention, and less to higher layers with focused patterns. Meanwhile, AdaKV~\citep{feng2024adakv} proposes a dynamic allocation framework that assigns varying cache budgets to different attention heads within a layer, based on the intra-layer attention distributions. In addition, to validate the Visual Head's effectiveness, we compare it with a Random Head baseline, where head scores are randomly initialized.
% In addition, to underscore the effectiveness of the Visual Head, we introduce a Random Head baseline for comparison. In this baseline, the only difference lies in the initialization of the scores for each head; they are randomly assigned rather than being derived from the Visual Head. This approach serves as a control that allows us to isolate and evaluate the specific contribution of the visual head component to the overall performance of the model.

\paragrapha{Benchmarks. }
To comprehensively assess the effectiveness of Visual Head in visual perception, we conduct evaluations on five widely used benchmarks covering both visual question answering (VQA) and image captioning tasks. Specifically, we utilize DocVQA~\citep{mathew2021docvqa}, OCRBench~\citep{liu2023ocrbench}, TextVQA~\citep{singh2019textvqa}, ChartQA~\citep{masry2022chartqa}, and TextCaps~\citep{sidorov2020textcaps}, which collectively encompass a diverse set of challenges, including document understanding, OCR-based question answering, chart interpretation, and text-based image captioning. Additionally, we also select the mainstream multiple-choice visual benchmarks, including MMBench~\citep{liu2023mmbench} and VQAv2~\citep{goyal2017vqav2} for a comprehensive evaluation. %Since our method applies compression after the prefilling stage, the first token is always accurate, making it infeasible to directly evaluate the model on these multiple-choice tasks. To address this issue, we slightly modified the prompt, requiring the model to answer in the format “The answer is \{your\_answer\}.” We then evaluated the responses using string matching, thereby circumventing the original problem. However, this requires the model to possess strong instruction-following capabilities, which the LLaVA series is unable to achieve. Therefore,  we only evaluated Qwen2-VL-7B-Instruct for the general visual benchmarks.

\subsection{Results on Multi-Modal Benchmarks}\label{subsec:exp_result}

\begin{table}[t]
    \centering
    \caption{\textbf{Average Number of Input Tokens.} We analyzed the average length of input tokens across various benchmarks. Considering that text instructions are typically very short(fewer than 50 tokens), so visual tokens constitute the majority of the input sequence, accounting for 90\% to 99\% of the input tokens.}
    \adjustbox{width=\linewidth}{
    \begin{tabular}{cccccc}
    \toprule
    Dataset           &  DocVQA & OCRBench & TextVQA & ChartQA & TextCaps \\
    \midrule
    LLaVA-Series      &  2433   &   1700   &  2376   &  2270   &   2376   \\
    Qwen2-VL-7B-Instruct &  4830   &   1245   &  1024   &   642   &   1024   \\
    \bottomrule
    \end{tabular}
    \label{tab:length}}
\end{table}

\paragrapha{Setups. }
To determine an appropriate budget allocation, we first measure the average length of input tokens for each benchmark, as reported in Table \ref{tab:length}. Given that text instructions typically consist of no more than 50 tokens, the majority of input tokens are attributed to visual tokens. Considering the varying input sequence lengths across different datasets, we select a range of each head's KV Cache budget for evaluation: \{64, 128, 256, 512, 1024, 2048\}. Since general visual benchmarks utilize lower image resolutions, we adjust the input token budget range correspondingly: \{48, 64, 96, 128, 256, 512\}. This allows us to systematically analyze the impact of different cache sizes on performance and efficiency across various benchmarks. 

\paragrapha{Results. }Fig.~\ref{fig:main_result} presents the evaluation results for three models and five benchmarks. Our experimental results demonstrate that our proposed method consistently outperforms baseline approaches, particularly under extreme cache budget constraints (e.g., 128 or 256). Under these conditions, our approach maintains performance levels close to those achieved with full cache utilization, significantly outperforming the competing baselines. For instance, on the TextVQA~\citep{singh2019textvqa} task using LLaVA-NeXT-Vicuna-7B, a KV Cache budget of 256—which constitutes only approximately 10.77\% of the average 2376 tokens—yields performance equivalent to the full-cache model, whereas AdaKV~\citep{feng2024adakv} and similar methods experience an accuracy drop of roughly 3\%. Similarly, on OCRBench~\citep{liu2023ocrbench}, LLaVA-NeXT-Mistral-7B demonstrates only a slight performance degradation at a KV Cache budget of 128 (about 7.5\% of the average 1700 tokens), in contrast to a decline exceeding 10\% observed with other methods. In addition, Qwen2-VL-7B-Instruct on DocVQA~\citep{mathew2021docvqa} maintains performance when operating with a KV Cache budget of 256 (merely 5.3\% of the average 4830 tokens), while alternative approaches suffer performance drops between 5\% and 17\%. These results validate the effectiveness of our method in VQA tasks. %On TextCaps~\citep{sidorov2020textcaps}, our method achieves the best performance, demonstrating that even with limited budgets, the visual head effectively attends to and retains the appropriate image information. For instance, LLaVA-NeXT-Vicuna-7B slightly outperforms the original baseline when using a KV budget of 256 and shows only a marginal decline compared to the baseline when the KV budget is reduced to 128 (which represents just 5.4\% of the original 2,376 tokens). Similar results are observed with Qwen2-VL-7B-Instruct. Although LLaVA-NeXT-Mistral-7B experiences a performance drop compared to baseline, it still significantly outperforms all the baselines.

\begin{figure}[t]
  \centering
  \includegraphics[width=\linewidth]{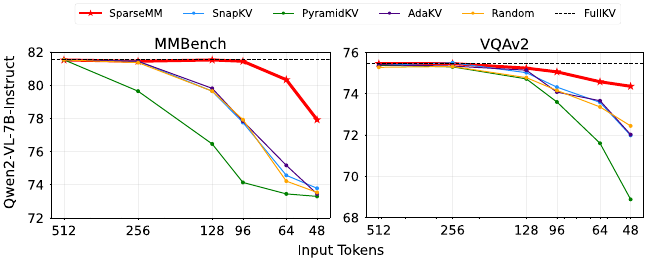}\vspace{-5pt}
  \caption{\textbf{Results on Multiple-choice Benchmarks.} We evaluate SparseMM and other baselines on multiple-choice visual benchmarks with Qwen2-VL-7B-Instruct as the backbone model. Our SparseMM consistently outperforms the other baselines.}
  \label{fig:general_visual_result}\vspace{-10pt}
\end{figure}

Fig.~\ref{fig:general_visual_result} presents the evaluation results on multiple-choice benchmarks. Our method demonstrates competitive performance on multi-choice benchmarks compared with existing baselines.  For instance, with only 96 token budget, our approach retains full performance on MMBench while experiencing only a minimal performance degradation ($<$1\%) on GQA and VQAv2.  These findings substantiate that our method can effectively recognize visual content while exhibiting strong generalizability towards diverse tasks.

Furthermore, our findings in Fig.~\ref{fig:main_result} indicate that the random head baseline consistently produces the poorest performance across nearly all experiments, whereas our method achieves superior outcomes by utilizing Visual Head. This pronounced contrast underscores the efficacy of our approach in accurately capturing the manner in which multi-modal language models attend to visual information. It is important to note that the performance of the random head method is comparable to that of SnapKV~\citep{li2024snapkv}, particularly in the case of the MHA model. This similarity is attributable to the fact that when the scores of all heads are randomly initialized, the cache budget allocated to each head is statistically equivalent, effectively causing the method to revert to the behavior observed with SnapKV~\citep{li2024snapkv}.

\subsection{Efficiency Evaluation}\label{subsec:exp_efficiency}

\paragrapha{Setup}
In this subsection, we evaluate the computational efficiency of our proposed method, which holds significant practical value. Accordingly, our efficiency tests are conducted across a range of input token lengths \{2K, 4K, 8K, 16K, 32K\}. For each experiment, the output sequence length was fixed at 100 tokens, with a KV Cache budget set to 256. We computed the average decoding latency and peak memory consumption for each configuration. Notably, all experiments are done using FlashAttention.

\begin{figure}[t]
  \centering
  \includegraphics[width=\linewidth]{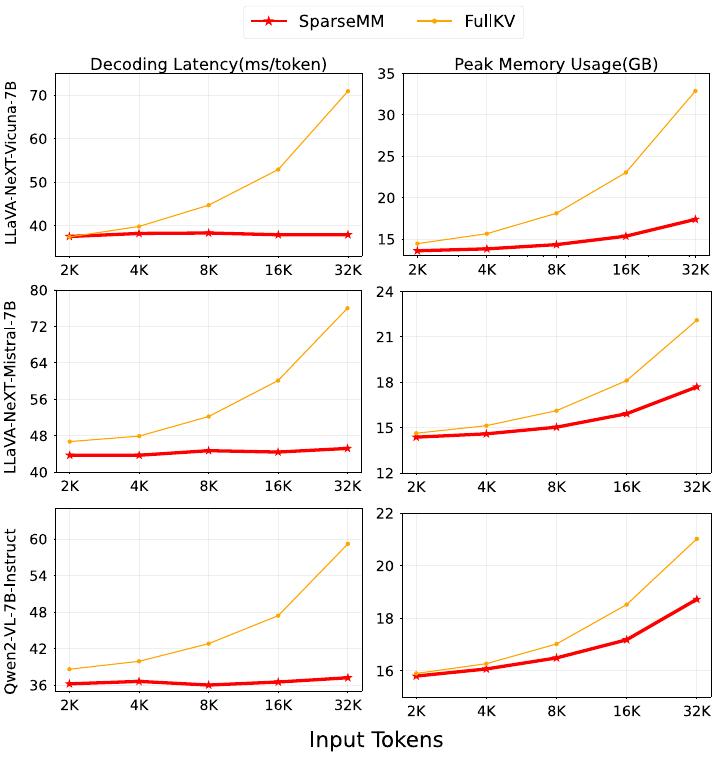}\vspace{-5pt}
  \caption{\textbf{Efficiency Evaluation for \emph{SparseMM}. } Benefiting from the reduction in KV cache, SparseMM can maintain nearly constant decoding latency, achieving up to a 50\% acceleration. Additionally, it effectively reduces peak memory usage.}
  \label{fig:efficiency}\vspace{-10pt}
\end{figure}

\paragrapha{Decoding Latency}
Fig.~\ref{fig:efficiency} illustrates that the reduction in KV Cache in our method substantially decreases the computational load during inference, thereby enhancing inference speed. For instance, when the input sequence length is 8K, the LLaVA-NeXT-Vicuna-7B model exhibits a speedup of $1.16\times$, while at a 32K input length, the speedup increases to $1.87\times$. These findings indicate that our approach significantly accelerates token generation, particularly in high-resolution or long video contexts.

\paragrapha{Memory Cost}
Our method also offers a marked reduction in peak memory usage, primarily by diminishing the memory footprint associated with the KV Cache. This reduction is especially pronounced in LLaVA-Series models. For example, with an input sequence length of 32K, LLaVA-NeXT-Vicuna-7B with full KV Cache requires 32.87 GB of memory, whereas our method reduces the requirement to 17.38 GB, thereby achieving an approximate 50\% reduction in memory overhead. It is noteworthy that even for the Qwen2-VL-7B-Instruct model, which employs an aggressive compression technique in its GQA framework, we can still reduce the cache by nearly 2GB with 32k inputs.

\subsection{Analysis}

\begin{figure}[t]
  \centering
  \includegraphics[width=\linewidth]{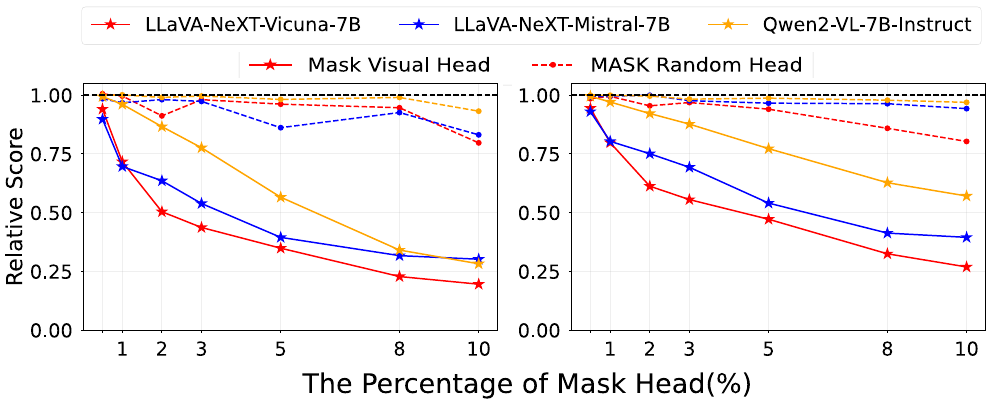}\vspace{-5pt}
  \caption{\textbf{Comparisons of Mmasking \emph{Visual Head} and Random Head. } The left figure is the result on OCRBench, and the right figure is the result on TextVQA. } %When masking only the top 3\% of visual heads with the highest scores, the model's performance dropped by half, while randomly masking heads resulted in almost no change in performance. The results indicate that visual heads constitute a small proportion of all heads while they are crucial for visual information perception.}
  \label{fig:mask}\vspace{-10pt}
\end{figure}

\paragrapha{Performance Influence of Visual Heads. }\label{subsec:exp_analysis}
To further elucidate the impact of visual heads on the visual perception capabilities of multimodal models, we conducted a series of masking experiments. In these experiments, we selectively masked a specific proportion of the visual heads and, for comparison, randomly masked an equivalent proportion of attention heads. The evaluation was performed on OCRBench~\citep{liu2023ocrbench} and TextVQA~\citep{singh2019textvqa}, with performance measured relative to the baseline unmasked model. The results, as illustrated in Fig.~\ref{fig:mask}, reveal that masking visual heads leads to a significant performance decline. In contrast, randomly masking the same proportion of attention heads produced a much smaller impact—for instance. These findings underscore the critical role that visual heads play in enabling MLLMs to effectively capture and process visual information. Moreover, masking the top 5\% of high-scoring visual heads causes a considerably greater performance loss than the additional impact of masking another 5\%, which highlights the sparse yet indispensable distribution of visual heads.

\begin{figure}[t]
  \centering
  \includegraphics[width=\linewidth]{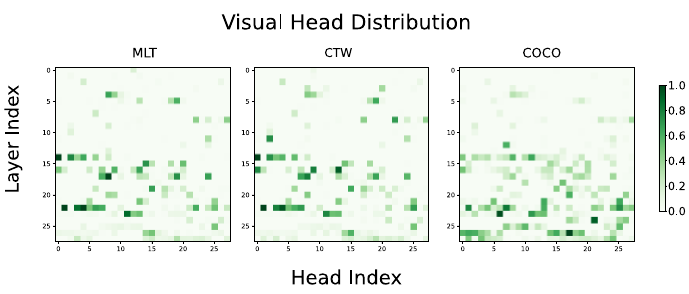}\vspace{-5pt}
  \caption{\textbf{Visualizations of \emph{Visual Heads} using Different Datasets.} We visualized the attention distribution identified on MLT, CTW, and COCO datasets.}
  \label{fig:coco_head}\vspace{-10pt}
\end{figure}

\begin{figure}[t]
  \centering
  \includegraphics[width=\linewidth]{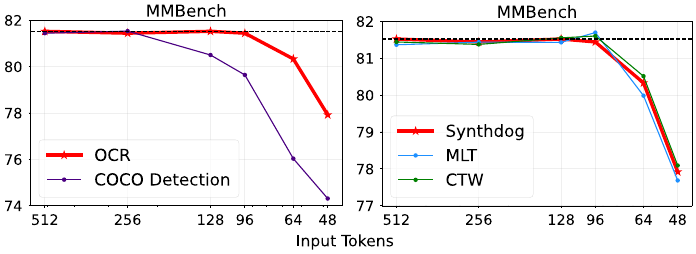}\vspace{-5pt}
  \caption{\textbf{Results with Different Visual Head Identification Approaches and Datasets.} We conduct an evaluation on visual heads identified on different datasets. The results on OCR datasets are similar and better than those on the detection dataset.}
  \label{fig:coco_mmbench}\vspace{-10pt}
\end{figure}

\paragrapha{Robustness of Visual Head Identification.} To evaluate the robustness of our visual head detecting approach, we show the distribution of the detected visual heads under different datasets and tasks in Fig.~\ref{fig:coco_head} and show the accuracy curve in Fig.~\ref{fig:coco_mmbench}. For the OCR task, we used the Multi-Lingual Text(MLT)~\citep{nayef2019mlt} and Chinese Text in the Wild(CTW)~\citep{yuan2019ctw} datasets. In addition, we consider the object detection task and choose the COCO dataset~\citep{lin2014coco}, where the model is required to identify objects present in the images. We then localized the visual heads based on the correspondence between the model’s answers and the relevant objects. As shown, the distribution of visual heads is relatively consistent across the OCR datasets, whereas there is greater variation on the COCO dataset. Moreover, experimental results demonstrate that the visual heads identified from OCR tasks are dataset-agnostic and exhibit strong generalizability, with better results than detection tasks. This is because OCR tasks establish an exact one-to-one mapping between the model’s output and the visual content, whereas the COCO task, which focuses on larger bounding boxes, introduces more noise and results in less robustness.

\paragrapha{Accuracy and Speed Trade-off.}
We compared the accuracy and speed of SparseMM with other baselines in Tab.~\ref{tab:speed}. We conducted an experiment on LLaVA-NeXT-Vicuna-7B with a budget of 256 KV Cache. With the support of FlashAttention, our decoding latency is comparable to that of other methods, significantly lower than the FullKV method. However, our method outperforms others in terms of performance under the same budget. This effectively demonstrates the efficacy of SparseMM based on visual heads in MLLMs.

\newcommand\cb[1]{\color{blue} #1}
\newcommand\cred[1]{\color{red} #1}
\begin{table}[t]
    \centering
    \caption{\textbf{Comparison of Accuracy-speed Trade-off among Different Methods. }We compare the speed of all methods under 256 KV Cache budget and 16K input tokens.}
    \adjustbox{width=\linewidth}{
    \begin{tabular}{ccccccc}
    \toprule
        Methods  &  DocVQA &  OCRBench & TextVQA & ChartQA & TextCaps & Latency(ms)\\
    \midrule
    FullKV &  0.68   &   0.52   &  0.65   &  0.55   &   0.73  & 52.9\\
    \midrule
    SparseMM &  \textbf{0.68}\cb{(-0.00)}   &  \textbf{0.52}\cb{(-0.00)}   &  \textbf{0.65}\cb{(-0.00)}   &  \textbf{0.54}\cb{(-0.01)}   &  \textbf{0.73}\cb{(-0.00)}  & 37.1\cred{(-30\%)}\\
    SnapKV &  0.64\cb{(-0.04)}  &   0.46\cb{(-0.06)}  &  0.62\cb{(-0.03)}  &   0.50 \cb{(-0.05)}  &   0.65\cb{(-0.08)}  & 35.3\cred{(-33\%)}\\
    PyramidKV &  0.65\cb{(-0.03)} &  0.48\cb{(-0.04)}  &  0.62\cb{(-0.03)}  &   0.53\cb{(-0.02)}   &   0.65\cb{(-0.08)} & \textbf{34.9}\cred{(-34\%)}\\
    AdaKV  &  0.65\cb{(-0.03)}  &   0.48\cb{(-0.04)}   &  0.62\cb{(-0.03)}   &   0.49\cb{(-0.06)}  &   0.66\cb{(-0.07)}  & 37.3\cred{(-29\%)}\\
    \bottomrule
    \end{tabular}
    \label{tab:speed}}
\end{table}

\begin{figure}[t]
  \centering
  \includegraphics[width=\linewidth]{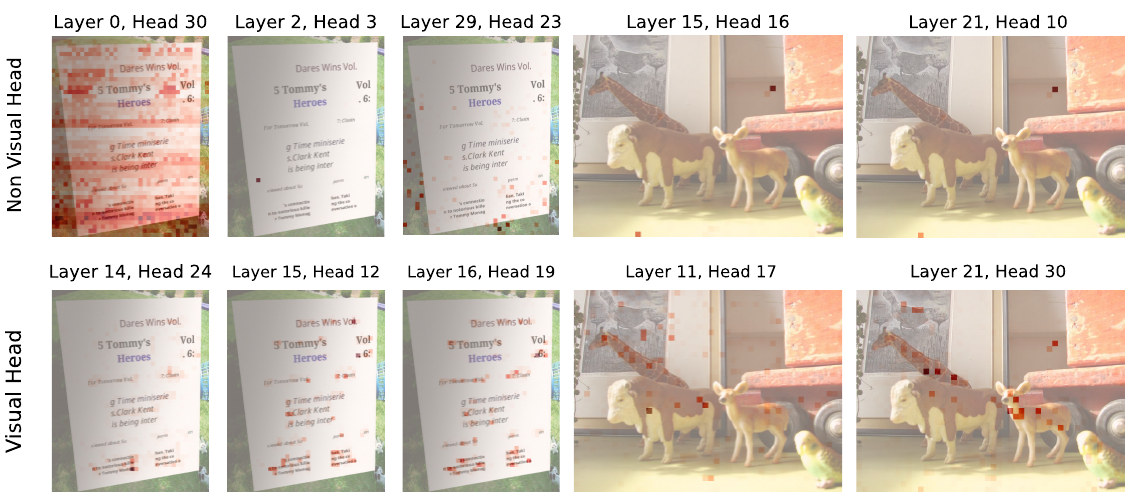}\vspace{-5pt}
  \caption{\textbf{Visualizations of \emph{Visual Heads}. } We visualized the attention distribution of several heads. The visual heads are able to accurately capture text or objects within the images, whereas the non-visual heads provide random results.}
  \label{fig:visual}\vspace{-10pt}
\end{figure}

\paragrapha{Visualization of Visual Heads. }\label{subsec:exp_visual}
To gain a more intuitive understanding of how visual heads process and interpret visual information, we conducted a visualization analysis of visual heads and non-visual heads on LLaVA-NeXT-Vicuna-7B. As illustrated in Fig.~\ref{fig:visual}, our observations indicate that non-visual heads often either neglect the image entirely or allocate a disproportionate amount of attention to visually insignificant regions. In contrast, visual heads accurately pinpoint regions of interest, allocating a substantial proportion of attention to these critical areas. 
% This precise focus explains why visual heads are particularly effective in capturing and encoding visual concepts. Moreover, these visualization results underscore the functional disparity between visual and non-visual heads and highlight the importance of dedicated visual attention mechanisms in enhancing the overall perceptual capabilities of multi-modal models.

\section{Conclusion}

In this paper, we present a systematic exploration of the visual processing characteristics inherent in MLLMs.    Our analysis reveals a critical sparsity phenomenon that only a small fraction of attention heads actively engage in visual understanding. Leveraging this insight, we propose SparseMM, a novel KV-Cache optimization framework that dynamically allocates asymmetric computation budgets to attention heads based on their visual relevance. SparseMM prioritizes preserving vision-critical information during decoding, thereby achieving a more balanced accuracy-efficiency trade-off. We hope this study inspires deeper investigations into the principles governing multimodal learning.

\appendix

\section*{Appendix}

\section{Implementation details}

\subsection{Implementation details about GQA}
The models LLaVA-NeXT-Mistral-7B, and Qwen2-VL-7B-Instruct are Grouped-Query Attention (GQA) models, which differ markedly from conventional multi-head attention (MHA) mechanisms in the computation of attention. In a GQA model, the query state is of shape $(bs, seq\_len,num\_query\_heads,hidden\_dim)$, while the key and value states, collectively forming the KV cache, are of shape $(bs,seq\_len,$ $num\_key\_value\_heads,hidden\_dim)$. During the attention calculation, the key and value states are repeated 
\[
\frac{num\_query\_heads}{num\_key\_value\_heads} = num\_key\_value\_group
\]
times, thereby restoring the setup analogous to MHA. Prior to computation, the sequence length dimension and the query head dimension are interchanged, resulting in an attention score tensor of shape $(bs, num\_query\_heads,seq\_len,seq\_len)$.
Subsequently, when this tensor is combined with the value states, the output is of shape $(bs, num\_query\_heads,seq\_len,hidden\_dim)$

From the above reasoning, it follows that we obtain a visual head score matrix with dimensions $(layers, num\_query\_head)$. This is the origin of the score distribution depicted in Fig. 2.

In practical scenarios involving the preservation of the Key-Value cache, each key-value head is associated with \(\text{num\_key\_value\_group}\) attention scores. The total attention score for a given head is computed as the sum of the scores of the corresponding group. This aggregate score is then employed for the allocation of the budget for the Key-Value cache.

\subsection{Details on Evaluation Metrics}

We adopt different evaluation metrics for different benchmarks. For the DocVQA~\citep{mathew2021docvqa} benchmark, we employ the ANLS metric. This metric evaluates the similarity between the predicted answer and the ground truth by normalizing the Levenshtein distance, thereby accommodating minor variations in format and phrasing while maintaining a robust assessment of answer quality. For the OCRBench~\citep{liu2023ocrbench}, TextVQA~\citep{singh2019textvqa}, MMBench~\citep{liu2023mmbench}, GQA~\citep{ainslie2023gqa} and VQAv2~\citep{goyal2017vqav2} benchmark, we use accuracy as the primary metric. For ChartQA~\citep{masry2022chartqa} benchmark, we utilize the relaxed accuracy metric. This measure provides partial credit for responses that are close to the ground truth, thereby offering a more nuanced perspective on model performance when outputs are not perfectly correct but still largely informative. Finally, for the TextCaps~\citep{sidorov2020textcaps} dataset, we adopt the CIDEr metric. CIDEr assesses the quality of generated captions by computing a weighted n-gram similarity between the candidate and reference captions.

\section{More Visualization}
\begin{figure}[t]
  \centering
  \includegraphics[width=\linewidth]{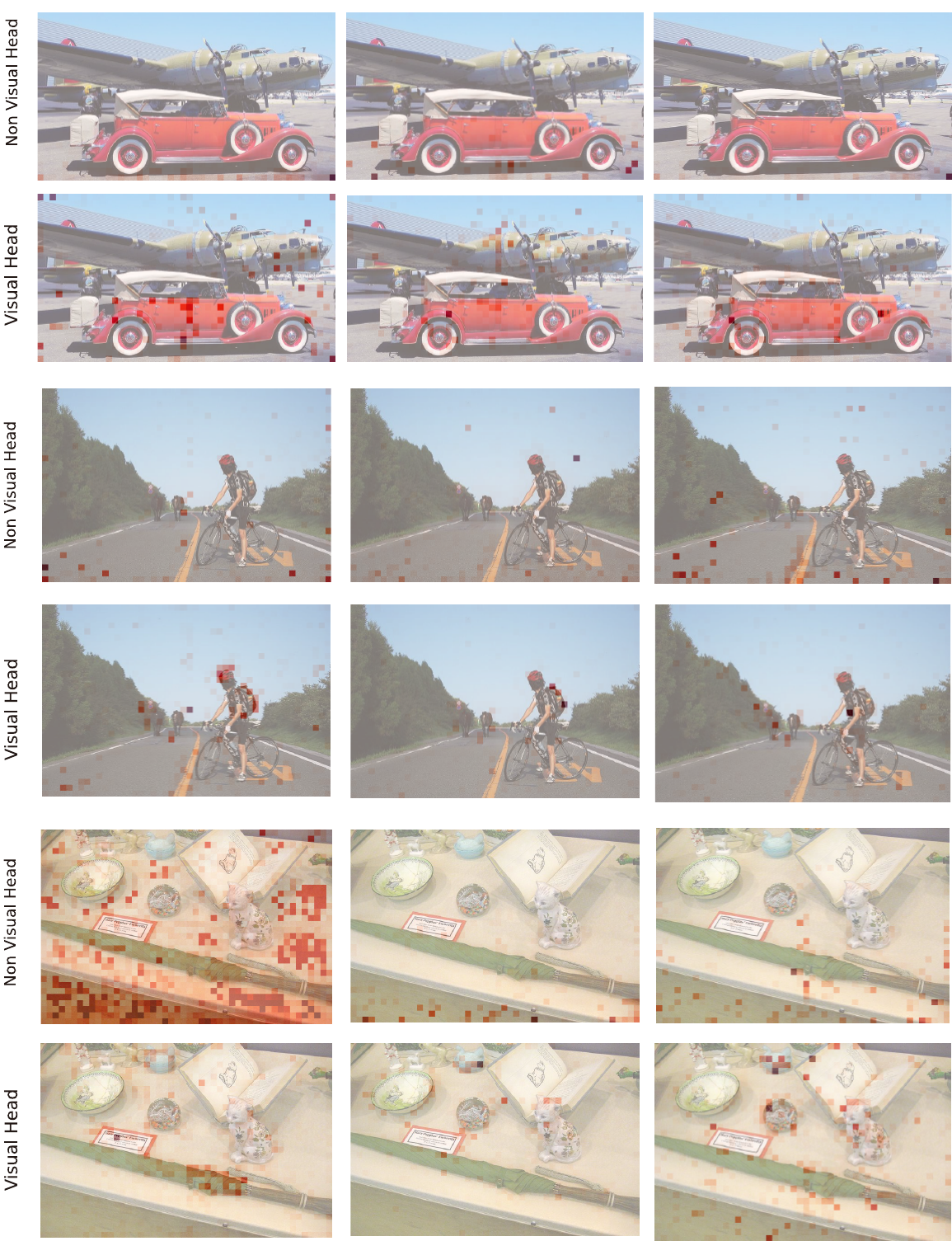}\vspace{-5pt}
  \caption{\textbf{More Visualization Results.} Visual heads are able to attend to the correct objects, whereas non-visual heads cannot.}
  \label{fig:supp}\vspace{-10pt}
\end{figure}

We conduct more visualization on the visual head in Fig.~\ref{fig:supp}. We use LLaVA-NeXT-Vicuna-7B model for the experiment.

\section{More Analysis}

\paragrapha{Ablations on Budget Allocation Ratios. }We conducted an ablation study on the hyperparameter $\rho$. This study evaluated the performance of three models on OCRBench, with a budget of 256. The results are presented in Tab.~\ref{tab:ablation}. For the LLaVA-NeXT-Vicuna-7B model, the ratio $\rho = 0.1$ achieved the highest performance score of 0.522, outperforming other ratios. Similarly, for the LLaVA-NeXT-Mistral-7B model, a ratio of 0.1 also resulted in a peak performance score of 0.519, which is significantly higher compared to the scores at other ratios. While the Qwen2-VL-Instruct model exhibited only a marginally higher score at $\rho = 0.1$ (0.812), this still represents the highest performance across all tested ratios. It is noteworthy that the Mistral model exhibits a significant performance drop at a ratio of $\rho=0$. This observation suggests that relying entirely on visual head score allocation of the cache budget can result in some heads being unable to attend to any preceding input information. Consequently, this underscores the necessity of assigning a Uniform-Based cache to each head. By ensuring that each head receives a guaranteed share of the cache resources, we can prevent such performance degradation and enhance the overall effectiveness of the model.

\begin{table}[t]
    \centering
    \caption{\textbf{Ablation on Budget Allocation Ratios.} We conducted an ablation study on the hyperparameter $\rho$ and the results indicated that the performance is optimal when the ratio is set to 0.1. Therefore, we use 0.1 as the default value in our experiments.}
    \adjustbox{width=\linewidth}{
    \begin{tabular}{ccccccccc}
    \toprule
    Ratio $\rho$           &  0 & 0.1 & 0.2 & 0.3 & 0.4 & 0.5 & 0.8 & 1.0\\
    \midrule
    LLaVA-NeXT-Vicuna-7B   & 0.507 & \textbf{0.522} & 0.520 & 0.520 & 0.516 & 0.515 & 0.510 & 0.460  \\
    LLaVA-NeXT-Mistral-7B  & 0.145 & \textbf{0.519} & 0.517 & 0.514 & 0.514 & 0.518 & 0.506 & 0.451 \\
    Qwen2-VL-7B-Instruct      & 0.809 & \textbf{0.812} & 0.811 & 0.808 & 0.807 & 0.804 & 0.789 & 0.775  \\
    \bottomrule
    \end{tabular}
    \label{tab:ablation}}
\end{table}

\paragrapha{Ablation on Cache Allocation Strategies.} We add an ablation study on Qwen2-VL-7B-Instruct to investigate the effectiveness of the three-part cache allocation mechanism. As shown in Tab.~\ref{tab:abaltion_cache}, using only Local-Window Cache limits context and causes larger drops with smaller budgets. Combining Local-Window and Uniform-Based Caches lacks head-level allocation and underperforms compared to our SparseMM. 

\begin{table}[t]
\centering
  \scriptsize
  \caption{\textbf{Ablation on Cache Allocation Strategies.} The results demonstrate that each of the three cache components plays an essential role and that none can be omitted without negatively impacting overall performance.}
    \label{tab:abaltion_cache}
    \adjustbox{width=\linewidth}{
    \begin{tabular}{ccccccccc}
    \toprule
    Local Window & Uniform-Based & Score-Preferred & \multicolumn{6}{c}{MMBench} \\
\cmidrule{4-9}  Cache  &  Cache & Cache & 512 & 256 & 128 & 96  & 64 & 48  \\

    \midrule
    \cmark & \xmark & \xmark & 81.3 & 80.5 & 77.3 & 73.6 & 70.5 & 67.2 \\
    \cmark & \cmark & \xmark & 81.5 & 81.4 & 79.3 & 77.6 & 74.6 & 73.9 \\
    \cmark & \cmark & \cmark & 81.5 & 81.4 & 81.5 & 81.4 & 80.3 & 77.9 \\
    \bottomrule
    \end{tabular}%
    }
\end{table}

\section{Numerical results}
We present the numerical results of our main experimental results for reference and further research.

\begin{table*}[htbp]
\centering
\scriptsize
\setlength{\tabcolsep}{2.5pt}
\caption{Numerical results of Fig.~\ref{fig:main_result}.}
\begin{tabular}{cc|cccccc|cccccc|cccccc}
    \toprule
        \multirow{2}{*}{Benchmark} 
        & \multirow{2}{*}{Method} 
        & \multicolumn{6}{c|}{LLaVA-NeXT-Vicuna-7B} 
        & \multicolumn{6}{c|}{LLaVA-NeXT-Mistral-7B} 
        & \multicolumn{6}{c}{Qwen2-VL-7B-Instruct} \\
        % \cmidrule(lr){3-8} \cmidrule(lr){9-14} \cmidrule(lr){15-20}
        & & 2048 & 1024 & 512 & 256 & 128 & 64 
          & 2048 & 1024 & 512 & 256 & 128 & 64 
          & 2048 & 1024 & 512 & 256 & 128 & 64 \\
    \midrule
    \multirow{5}{*}{DocVQA} 
    & SparseMM  
          & 0.6841 & 0.6837 & 0.6811 & 0.6784 & 0.6677 & 0.6377
          & 0.6310 & 0.6272 & 0.6227 & 0.6163 & 0.6082 & 0.5756
          & 0.9394 & 0.9392 & 0.9394 & 0.9345 & 0.9154 & 0.8493 \\
    &SnapKV    
          & 0.6845 & 0.6807 & 0.6709 & 0.6430 & 0.5906 & 0.4977
          & 0.6365 & 0.6306 & 0.6215 & 0.5971 & 0.5519 & 0.4726 
          & 0.9384 & 0.9340 & 0.9194 & 0.8798 & 0.8012 & 0.6652 \\
    &PyramidKV   
          & 0.6843 & 0.6812 & 0.6714 & 0.6494 & 0.6030 & 0.4901
          & 0.6363 & 0.6225 & 0.6076 & 0.5818 & 0.5425 & 0.4351
          & 0.9391 & 0.9343 & 0.8816 & 0.8180 & 0.7394 & 0.5990 \\
    &AdaKV     
          & 0.6839 & 0.6823 & 0.6753 & 0.6526 & 0.6064 & 0.5411
          & 0.6358 & 0.6299 & 0.6174 & 0.5957 & 0.5592 & 0.4872
          & 0.9392 & 0.9330 & 0.9201 & 0.8841 & 0.8121 & 0.6847 \\
    &Random     
          & 0.6834 & 0.6791 & 0.6646 & 0.6340 & 0.5816 & 0.4868
          & 0.6326 & 0.6217 & 0.5971 & 0.5592 & 0.4973 & 0.4206
          & 0.9275 & 0.9015 & 0.8534 & 0.7681 & 0.6963 & 0.5102 \\
    \midrule
    \multirow{5}{*}{OCRBench} 
    & SparseMM  
          & 0.519 & 0.522 & 0.528 & 0.523 & 0.501 & 0.478
          & 0.523 & 0.518 & 0.512 & 0.519 & 0.507 & 0.462
          & 0.821 & 0.822 & 0.821 & 0.812 & 0.795 & 0.743 \\
    &SnapKV    
          & 0.525 & 0.518 & 0.510 & 0.461 & 0.412 & 0.340
          & 0.529 & 0.517 & 0.500 & 0.450 & 0.390 & 0.319
          & 0.819 & 0.813 & 0.801 & 0.773 & 0.719 & 0.624 \\
    &PyramidKV   
          & 0.525 & 0.524 & 0.502 & 0.476 & 0.409 & 0.312
          & 0.528 & 0.512 & 0.489 & 0.440 & 0.394 & 0.290 
          & 0.820 & 0.814 & 0.776 & 0.739 & 0.682 & 0.563 \\
    &AdaKV     
          & 0.524 & 0.517 & 0.508 & 0.484 & 0.430 & 0.351
          & 0.529 & 0.520 & 0.502 & 0.451 & 0.404 & 0.328
          & 0.819 & 0.812 & 0.796 & 0.778 & 0.710 & 0.621 \\
    &Random     
          & 0.523 & 0.516 & 0.500 & 0.458 & 0.397 & 0.320
          & 0.521 & 0.507 & 0.451 & 0.399 & 0.354 & 0.261
          & 0.811 & 0.794 & 0.760 & 0.704 & 0.633 & 0.489 \\
    \midrule
    \multirow{5}{*}{TextVQA} 
    & SparseMM  
          & 0.6499 & 0.6492 & 0.6474 & 0.6470 & 0.6417 & 0.6312
          & 0.6555 & 0.6547 & 0.6531 & 0.6505 & 0.6474 & 0.6281
          & 0.8213 & 0.8215 & 0.8203 & 0.8218 & 0.8164 & 0.7719 \\
    &SnapKV    
          & 0.6488 & 0.6474 & 0.6408 & 0.6229 & 0.6010 & 0.5616
          & 0.6565 & 0.6541 & 0.6503 & 0.6345 & 0.6103 & 0.5712
          & 0.8213 & 0.8212 & 0.8204 & 0.8031 & 0.7746 & 0.6990 \\
    &PyramidKV   
          & 0.6487 & 0.6483 & 0.6410 & 0.6277 & 0.6040 & 0.5502
          & 0.6566 & 0.6490 & 0.6430 & 0.6285 & 0.6088 & 0.5467
          & 0.8218 & 0.8218 & 0.8076 & 0.7774 & 0.7440 & 0.6547 \\
    &AdaKV     
          & 0.6482 & 0.6486 & 0.6429 & 0.6199 & 0.5988 & 0.5609
          & 0.6566 & 0.6530 & 0.6464 & 0.6289 & 0.6049 & 0.5685
          & 0.8213 & 0.8212 & 0.8185 & 0.7985 & 0.7695 & 0.7025 \\
    &Random     
          & 0.6478 & 0.6438 & 0.6373 & 0.6235 & 0.6011 & 0.5653
          & 0.6536 & 0.6494 & 0.6358 & 0.6134 & 0.5822 & 0.5400
          & 0.8202 & 0.8166 & 0.7943 & 0.7601 & 0.6955 & 0.5852 \\
    \midrule
    \multirow{5}{*}{ChartQA} 
    & SparseMM  
          & 0.5480 & 0.5452 & 0.5488 & 0.5392 & 0.5380 & 0.5276
          & 0.5280 & 0.5216 & 0.5236 & 0.5188 & 0.5116 & 0.4888
          & 0.8152 & 0.8152 & 0.8128 & 0.8160 & 0.8152 & 0.8016 \\
    &SnapKV    
          & 0.5480 & 0.5536 & 0.5416 & 0.5000 & 0.4527 & 0.4304
          & 0.5288 & 0.5236 & 0.5164 & 0.5016 & 0.4752 & 0.4272
          & 0.8140 & 0.8144 & 0.8144 & 0.8128 & 0.7964 & 0.7552 \\
    &PyramidKV   
          & 0.5488 & 0.5536 & 0.5496 & 0.5304 & 0.4716 & 0.4100
          & 0.5272 & 0.5228 & 0.5080 & 0.4920 & 0.4708 & 0.4068
          & 0.8140 & 0.8144 & 0.8144 & 0.8088 & 0.7924 & 0.7332 \\
    &AdaKV     
          & 0.5492 & 0.5540 & 0.5480 & 0.4912 & 0.4576 & 0.4384
          & 0.5292 & 0.5224 & 0.5156 & 0.5044 & 0.4780 & 0.4460
          & 0.8152 & 0.8156 & 0.8140 & 0.8080 & 0.7964 & 0.7592 \\
    &Random     
          & 0.5480 & 0.5476 & 0.5424 & 0.5304 & 0.4936 & 0.4372
          & 0.5272 & 0.5152 & 0.5060 & 0.4764 & 0.4428 & 0.3944
          & 0.8152 & 0.8152 & 0.8060 & 0.7876 & 0.7500 & 0.6696 \\
    \midrule
    \multirow{5}{*}{TextCaps} 
    & SparseMM  
          & 0.7320 & 0.7309 & 0.7334 & 0.7284 & 0.7071 & 0.5992
          & 0.7067 & 0.7054 & 0.6896 & 0.6795 & 0.6339 & 0.5238
          & 1.4697 & 1.4744 & 1.4919 & 1.4915 & 1.4299 & 1.0431 \\
    &SnapKV    
          & 0.7226 & 0.7167 & 0.6969 & 0.6495 & 0.5642 & 0.4431
          & 0.7070 & 0.6969 & 0.6970 & 0.6504 & 0.5579 & 0.4436
          & 1.4677 & 1.4744 & 1.4695 & 1.3598 & 1.1424 & 0.7940 \\
    &PyramidKV   
          & 0.7237 & 0.7254 & 0.6953 & 0.6491 & 0.5745 & 0.4164
          & 0.7061 & 0.6828 & 0.6592 & 0.6230 & 0.5495 & 0.4062
          & 1.4694 & 1.4680 & 1.2745 & 1.1151 & 0.9536 & 0.5669 \\
    &AdaKV     
          & 0.7263 & 0.7273 & 0.7039 & 0.6598 & 0.5923 & 0.4727
          & 0.7037 & 0.6953 & 0.6850 & 0.6459 & 0.5664 & 0.4400
          & 1.4690 & 1.4650 & 1.4631 & 1.3445 & 1.1461 & 0.8133 \\
    &Random     
          & 0.7297 & 0.7219 & 0.6803 & 0.6268 & 0.5355 & 0.4356
          & 0.7065 & 0.6980 & 0.6882 & 0.6472 & 0.5512 & 0.4368
          & 1.4690 & 1.4727 & 1.4812 & 1.3824 & 1.1627 & 0.8116 \\
    \bottomrule
\end{tabular}
\label{tab:all_results}
\end{table*}

\begin{table*}[htbp]
    \centering
    \caption{Numerical results of Fig.~\ref{fig:general_visual_result}.}
    \adjustbox{width=\textwidth}{
    \begin{tabular}{c|cccccc|cccccc|cccccc}
        \toprule
        \multirow{2}{*}{Method} 
          & \multicolumn{6}{c|}{MMBench} 
          & \multicolumn{6}{c|}{GQA} 
          & \multicolumn{6}{c}{VQAv2} \\
        & 512 & 256 & 128 & 96 & 64 & 48 
        & 512 & 256 & 128 & 96 & 64 & 48 
        & 512 & 256 & 128 & 96 & 64 & 48 \\
        \midrule
        SparseMM  
          & 81.52 & 81.44 & 81.52 & 81.44 & 80.33 & 77.92
          & 64.51 & 64.52 & 64.20 & 63.66 & 62.48 & 60.88
          & 75.46 & 75.46 & 75.24 & 75.06 & 74.58 & 74.36 \\
        SnapKV    
          & 81.52 & 81.44 & 79.64 & 77.75 & 74.57 & 73.79
          & 64.53 & 64.51 & 63.77 & 62.38 & 60.82 & 59.19
          & 75.38 & 75.50 & 75.02 & 74.32 & 73.58 & 71.98 \\
        PyramidKV 
          & 81.53 & 79.64 & 76.46 & 74.14 & 73.45 & 73.30
          & 63.80 & 63.47 & 62.05 & 60.65 & 59.41 & 59.37
          & 75.38 & 75.30 & 74.72 & 73.60 & 71.60 & 68.88 \\
        AdaKV     
          & 81.52 & 81.44 & 79.81 & 77.83 & 75.17 & 73.45
          & 64.52 & 64.65 & 63.52 & 62.55 & 61.59 & 59.20
          & 75.40 & 75.34 & 75.14 & 74.08 & 73.66 & 72.02 \\
        Random    
          & 81.52 & 81.36 & 79.64 & 77.92 & 74.22 & 73.54
          & 64.51 & 64.38 & 63.87 & 62.60 & 61.00 & 59.39
          & 75.28 & 75.32 & 74.78 & 74.16 & 73.36 & 72.44 \\
        \bottomrule
    \end{tabular}
    }
    \label{tab:qwen_combined}
\end{table*}

{
    \small
    \bibliographystyle{ieeenat_fullname}
    \bibliography{main}
}

\end{document}